\documentclass[twocolumn]{article}
\usepackage{natbib}
\usepackage{amsmath,amssymb,amsfonts}
\usepackage{algorithm}
\usepackage{algpseudocode}
\usepackage{graphicx}
\usepackage{textcomp}
\usepackage{xcolor}
\usepackage{natbib}
\usepackage{booktabs}
\usepackage{multirow}
\usepackage{url}
\usepackage{hyperref}
\usepackage{fancyhdr}
\newcommand{\tableref}[1]{Table~\ref{table:#1}}
\newcommand{\figref}[1]{Figure~\ref{fig:#1}}

\newcommand{\sectref}[1]{Section~\ref{sec:#1}}

\begin{document}

\title{Benchmarking Artificial Intelligence Models for Daily Coastal Hypoxia Forecasting

}

\author{
\small
Magesh Rajasekaran$^{1,2}$,
Md Saiful Sajol$^{3}$,
Chris Alvin$^{4}$,
Supratik Mukhopadhyay$^{1,2}$,
Yanda Ou$^{2,5}$,
Z.\ George Xue$^{2,5}$ \\[1ex]
\small
$^{1}$Department of Environmental Sciences, Louisiana State University, Baton Rouge, LA, USA\\
\small
$^{2}$Center for Computation and Technology, Louisiana State University, Baton Rouge, LA, USA\\
\small
$^{3}$Department of Computer Science, Louisiana State University, Baton Rouge, LA, USA\\
\small
$^{4}$Department of Computer Science, Furman University, Greenville, SC, USA\\
\small
$^{5}$Department of Oceanography and Coastal Sciences, Louisiana State University Baton Rouge, LA, USA
}

\maketitle

\thispagestyle{fancy}

\begin{abstract}
Coastal hypoxia, especially in the northern part of Gulf of Mexico, presents a persistent ecological and economic concern.
Seasonal models offer coarse forecasts that miss the fine-scale variability needed for daily, responsive ecosystem management.
We present study that compares four deep learning architectures for daily hypoxia classification: Bidirectional Long Short-Term Memory (BiLSTM), Medformer (Medical Transformer), Spatio-Temporal Transformer (ST-Transformer), and Temporal Convolutional Network (TCN).
We trained our models with twelve years of daily hindcast data from 2009-2020 
Our training data consists of 2009-2020 hindcast data from a coupled hydrodynamic-biogeochemical model.
Similarly, we use hindcast data from 2020 through 2024 as a test data.
We constructed classification models incorporating water column stratification, sediment oxygen consumption, and temperature-dependent decomposition rates.
We evaluated each architectures using the same data preprocessing, input/output formulation, and validation protocols.
Each model achieved high classification accuracy and strong discriminative ability with ST-Transformer achieving the highest performance across all metrics and tests periods (AUC-ROC: 0.982-0.992).
We also employed McNemar's method to identify statistically significant differences in model predictions.
Our contribution is a reproducible framework for operational real-time hypoxia prediction that can support broader efforts in the environmental and ocean modeling systems community and in ecosystem resilience.
The source code is available\footnote{https://github.com/rmagesh148/hypoxia-ai/}.
\end{abstract}

Deep learning, Time-series analysis, Environmental monitoring, Neural networks, Hypoxia forecasting, Louisiana-Texas shelf

%
%
%%%%%%%%%%%%%%%%%%%%%%%%%%%%%%%%%
%
%

\section{Introduction}

\begin{figure}[t!]
\centering
\includegraphics[trim=5 10 0 0, clip, width=0.75\columnwidth]{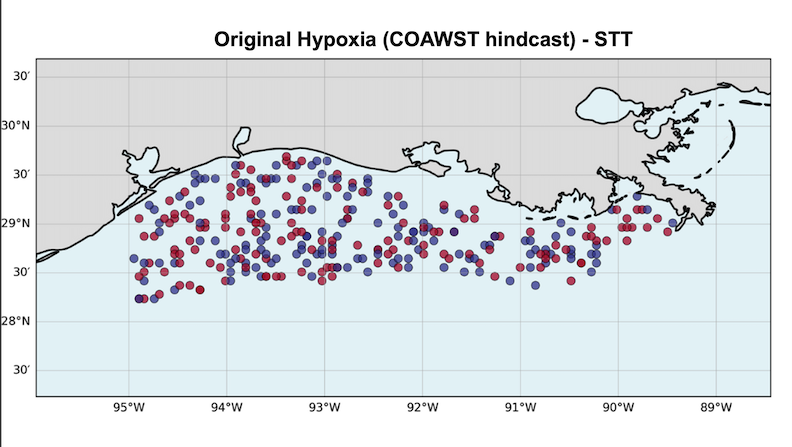}
\caption{The spatial domain of the hindcast model used in this study: the Louisiana-Texas shelf.
This image shows the random samples of COAWST hindcast hypoxia (Blue) and normoxia (Red) for the month of August 2020.}
\label{fig:spatial-domain}
\end{figure}

\begin{figure}[t!]
\centering
\includegraphics[width=0.55\columnwidth]{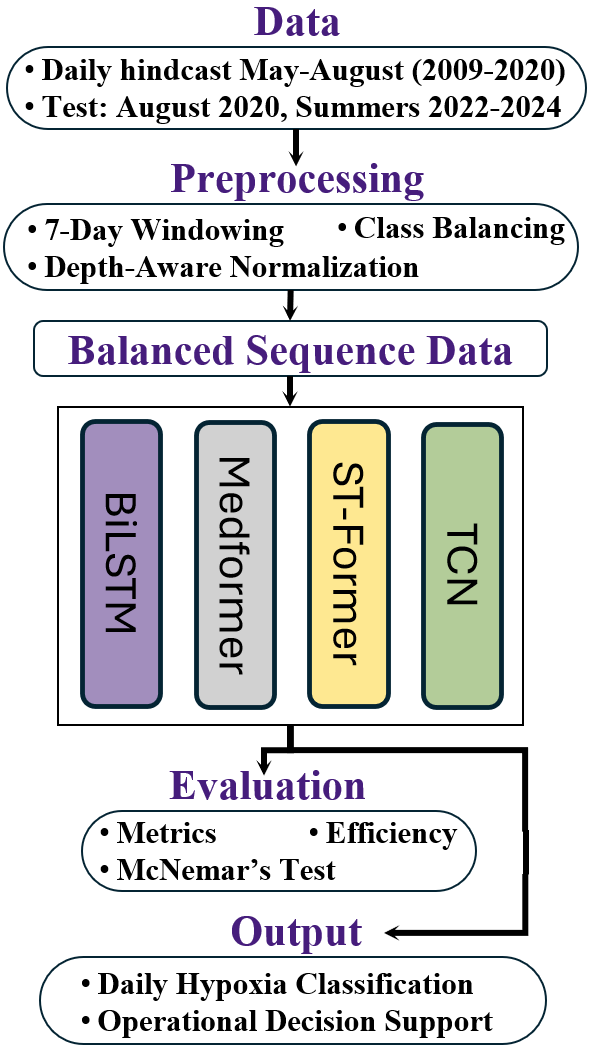}
\caption{Conceptual workflow for daily coastal hypoxia forecasting.}
\label{fig:pipeline}
\end{figure}

Dissolved oxygen concentrations below 2.0 mg/L is referred to as \emph{coastal hypoxia} and it creates serious problems for marine ecosystems worldwide.
In particular, the northern region of the Gulf of Mexico suffers from recurring seasonal hypoxic events that create ``dead zones'' that devastate marine life, disrupt fisheries, and impose real economic costs on coastal communities \cite{noaa2024deadzone}.
The Louisiana-Texas shelf (\figref{spatial-domain}) has some of the most severe and persistent hypoxic conditions globally, driven by interactions between nutrient loading from the Mississippi River, water column stratification, and processes involving sediment oxygen demand  \cite{ou-hydrodynamic}.

Current systems for operational forecasting in the Gulf use seasonal statistical models.
Temporally, monthly, or seasonal predictions are too coarse to provide useful forecasts.
These approaches miss the fine-scale temporal variability that is critical for responsive ecosystem management.
% MR Hypoxia events develop rapidly on daily timescales in response to wind events, changes in stratification, and river discharge variations.
Since these processes operate on timescales much shorter than seasonal cycles, greater temporal resolution in predictions is essential.

Daily resolution is required for operational forecasting; deep learning models can exploit temporal patterns in oceanographic time-series data to achieve this level of granularity.
Sequence modeling architectures such as recurrent neural networks~\cite{BiLSTM}, convolutional networks~\cite{tcn}, and transformer-based models~\cite{stformer}, have shown success in capturing complex temporal dependencies in different domains.
% MR However, their performance for coastal hypoxia prediction remains largely unexplored, particularly under controlled experimental conditions.

This paper addresses the need for daily hypoxia forecasting by comparing four deep learning architectures within a unified experimental framework (see \figref{pipeline}).
We formulate hypoxia prediction as a sequence-to-one classification task.
Our models are trained to learn how environmental conditions lead to oxygen depletion.
To this end, we use twelve years of daily hindcast data from a coupled hydrodynamic-biogeochemical model \cite{COAWST}.
We preprocess the daily hindcast data by (1) encoding temporal cycles, (2) normalizing, (3) creating sequences, and (4) addressing class imbalance between rare hypoxic events and the more common normoxic events.
The architectures we evaluate span the spectrum of modern sequence modeling approaches: BiLSTM for capturing bidirectional temporal dependencies, TCN for efficient parallel processing of temporal sequences, Medformer for multi-scale temporal pattern recognition, and ST-Transformer for joint spatial-temporal modeling.
We evaluate each model using the same framework to ensure statistical rigor in our comparisons: preprocessing, input/output formulation, and validation protocols.

Our contributions are as follows.
We benchmark temporal deep learning architectures for Gulf hypoxia prediction under identical experimental conditions, allowing direct performance comparison.
Second, we apply McNemar's statistical significance test to quantify differences in model predictions; this is notable as environmental AI studies often do not perform such an analysis.
Third, we evaluate transformer variants (Medformer and spatio-temporal attention models) that have yet to be tested for coastal hypoxia prediction.

%
%
%%%%%%%%%%%%%%%%%%%%%%%%%%%%%%%%%
%
%
\section{Task Overview}
\label{sec:overview}

Our goal is to develop models that support daily operational forecasting of hypoxic conditions.
We therefore formulate coastal hypoxia prediction as a binary classification problem to distinguish between hypoxic (dissolved oxygen $\leq 2.0$ mg/L) and normoxic conditions in bottom waters of the Louisiana-Texas shelf.
We do so using a sequence-to-one prediction framework that captures temporal dependencies in oceanographic processes.
Our input features include environmental time series that influence oxygen depletion in coastal marine ecosystems.

Daily hypoxia forecasts can provide early warnings of hypoxic events for fishery managers and researchers in the field.
We focus on metrics suitable for operational use: F1-score for handling class imbalance, AUC-ROC for ranking ability, and optimized accuracy thresholds to support decision-making.
Our goal is to maximize detection of hypoxic events while mitigating false alarm rates.

%
%%%%%%%%%%%%%%%%%%%%%%%%%
\section{Dataset and Data Preparation}
\label{sec:data}

% We describe the oceanographic hindcast dataset and the pipeline we used to prepare the data for learning.

%
%%%%%%%%%%%%%%%%%%%%%%%%%
\subsection{The Data}
We use a hindcast dataset from the Louisiana-Texas shelf covering May through August of each year from 2009 to 2020 \cite{ou2024forecasting,Ou2025} with further hindcast data covering 2022 through 2024.
The spatiotemporal records originate from Coupled Ocean-Atmosphere-Wave-Sediment Transport (COAWST) \cite{COAWST}, a modeling pipeline tailored for the Gulf of Mexico.
COAWST integrates the Regional Ocean Modeling System (ROMS) \cite{ROMS} with a revised  version of the North Pacific Ecosystem Model for Understanding Regional Oceanography (NEMURO) \cite{NEMURO}.

In total, our 2009-2020 dataset contains $1471$ summer records having $25 km^2$ spatial resolution organized as 2-dimensional matrices.
We include three key input variables known to drive coastal hypoxia.
Potential energy anomaly (PEA) measures water column stratification \cite{ou-hydrodynamic}; higher values indicate stronger stratification that reduces oxygen supply to bottom waters.
The rate of sediment oxygen consumption ($SOC$) \cite{SOC} and the temperature-dependent decomposition rate of organic matter ($DCP_{Temp}$) \cite{DCPTemp} represent two more input variables.
Both variables contribute to oxygen consumption in sediments since they influence the decomposition of organic matter within sediment layers.
Each of these variables are provided as multivariate time series spanning $7$-day windows to capture temporal dependencies inherent in oceanographic processes.

Our models classify conditions as hypoxic (bottom-dissolved oxygen concentration is $< 2.0 mg/L$) or normoxic.

%
%%%%%%%%%%%%%%%%%%%%%%%%%
\subsection{Data Preparation Pipeline}
\label{sec:pipeline}

\begin{figure}[t!]
\centering
\includegraphics[width=0.75\columnwidth]{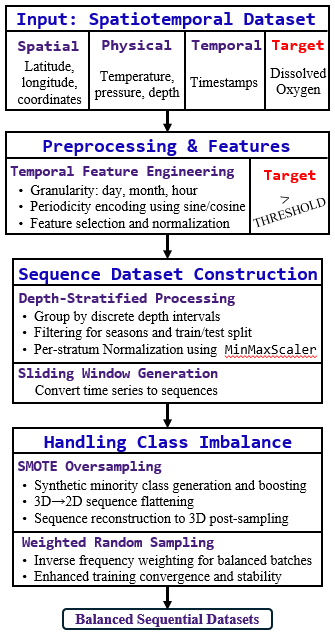}
\caption{The data preparation workflow for our Deep Learning pipeline for time series for classification of dissolved oxygen in the Louisiana-Texas shelf.}
\label{fig:data-preparation}
\end{figure}

Our data preparation pipeline (\figref{data-preparation}) converts raw hindcast data into sequences suitable for training and testing.

\textbf{Data input.}
We accept the hindcast matrices and organize them for sequential processing.
This involves extracting the temporal coordinates from each record to enable subsequent chronological ordering and feature engineering operations.

\textbf{Preprocessing and feature engineering.}
We transform the continuous oxygen measurements into binary classification targets and augment the temporal information through cyclical encoding.
That is, for each time-scale (day of year, month, hour), we extract and convert using sine and cosine transformations to preserve seasonal and diurnal periodicity.
This encoding ensures that temporal boundaries (e.g., December to January transitions, or \texttt{23:59} to \texttt{00:00} hour changes) are represented as continuous rather than discrete jumps, allowing a model to properly capture the circular nature of time.
Last, all matrices undergo minimum-maximum normalization, and any spatial regions corresponding to land are masked using zeroes.

\textbf{Building the Sequence Dataset.}
We then partition the time series data by depth to account for distinct oceanographic characteristics at each level.
We split the data temporally selecting August 2020 and data from 2022-2024 as test sets; all other months from 2009 to 2020 constitute the training set.
This ensures model evaluation preserves temporal integrity by relying only on `future' data.

To prepare the data for sequence modeling, we segment the time series into overlapping sequences using a sliding window approach.
This creates three-dimensional tensors with dimensions for samples, time steps, and features.
We then normalize each depth interval independently to preserve depth-specific oceanographic patterns.

\textbf{Mitigating imbalance of hypoxic events.}
Hypoxic events are relatively rare.
To mitigate the class imbalance in the dataset, we apply SMOTE oversampling \cite{chawla2002smote}, which has been successfuly applied to time-series classification tasks.
SMOTE requires 2D input, so we flatten each sequence by concatenating time steps and features into a single vector before generating synthetic samples.
SMOTE generates synthetic minority class samples by linearly interpolating between existing samples and their nearest neighbors:

\begin{equation}
x_{\text{new}} = x + \mu \cdot (x_{nn} - x), \quad \mu \sim U(0,1)
\end{equation}

\noindent with $x$ being a sample from a minority class, $x_{nn}$ being one of the $k$ nearest neighbors of $x$  belonging to the same minority class, and $\mu$ being a random number uniformly drawn from $[0,1]$.
After oversampling, the sequences are reshaped to their original 3D temporal structure. 
We combine SMOTE with weighted random sampling during training (\sectref{implementation}).
SMOTE addresses imbalance at the dataset level by increasing minority class representation while
weighted random sampling addresses imbalance at the batch level by ensuring balanced batch composition.
% We also use weighted random sampling when we generate batches, where samples are drawn based on inverse class frequency.
The resulting training data ensures that rare hypoxic cases are adequately represented.

%
%
%%%%%%%%%%%%%%%%%%%%%%%%%%%%%%%%%
%
%
\section{Architectures}
\label{sec:architecture}

% \begin{figure*}[t!]
% \centering
% \includegraphics[width=0.95\columnwidth]{figures/placeholder-architectures.png}
% \caption{\todo{Placeholder: architecture of the four models?}}
% \label{fig:architectures}
% \end{figure*}

As described in \sectref{data}, our dataset consists of multivariate environmental observations sampled daily from 2009-2020 and in summers of 2022-2024.
Hypoxia onset, persistence, and dissipation are governed by bottom-water oxygen levels \cite{noaa2024deadzone,yu2015physical,matli2024trends}.
These factors are modulated by wind events, stratification, and river-plume dynamics that operate on shorter timescales than seasonal or interannual cycles alone.
Sequence models are well-suited for these data because they capture temporal patterns and interactions among variables.
In this work, we consider four architectures for daily hypoxia classification: \emph{Bidirectional LSTM (BiLSTM)} \cite{schuster1997bidirectional,BiLSTM}, \emph{Temporal Convolutional Network (TCN)} \cite{tcn}, \emph{Medformer} \cite{wang2024medformer}, and \emph{Spatio-Temporal Transformer (ST-Transformer)} \cite{stformer}.

%
%%%%%%%%%%%%%%%%%%%%%%%%%
\textbf{Bidirectional LSTM.}
BiLSTM augments the standard LSTM with two parallel layers: one forward, one backward.
This bidirectional structure captures dependencies from both past and future within the observation window, which is useful when outcomes depend on cumulative conditions rather than isolated states.
The forward layer processes a sequence from past to present, with the backward layer processing  from future to past.
The final representation combines both temporal perspectives:

\begin{align}
\overrightarrow{s}_t &= \text{LSTM}_{f}(u_t, \overrightarrow{s}_{t-1}) ~ \text{\cite{BiLSTM}}\\
\overleftarrow{h}_t &= \text{LSTM}_{b}(u_t, \overleftarrow{s}_{t+1}) ~ \text{\cite{BiLSTM}}\\
s_t &= [\overrightarrow{s}_t ; \overleftarrow{s}_t]
\end{align}

\noindent where $\overrightarrow{s}_t$ and $\overleftarrow{s}_t$ are the hidden states from the respective forward and backward passes, and $s_t$ is their concatenation and $u_t$ is the $t$th input in the sequence.
The implementation uses two hidden layers, each containing 120 memory units.
Dropout regularization of 30\% is applied between layers to account for  overfitting, while the final classification layer maps the temporal representations to binary hypoxia predictions.

%
%%%%%%%%%%%%%%%%%%%%%%%%%
\textbf{Temporal Convolutional Network}
Recurrent architectures have long been the default for time-series modeling, yet they introduce challenges such as vanishing gradients and slow, sequential computation.
TCN replaces recurrence with dilated causal convolutions, preserving temporal order while enabling parallel computation.
By stacking convolutional layers with progressively increasing dilation rates across three layers, TCNs can capture long-range temporal dependencies through large receptive fields without requiring deep or computationally intensive architectures.
Temporal integrity is maintained since the causal structure depends only on past and current signals.
Each layer in the network applies dilated convolutions with exponentially increasing dilation rates:

\begin{align}
h_t^{(l)} &= \sigma \!\left( \sum_{k=0}^{K-1} W_k^{(l)} \, h_{t-d \cdot k}^{(l-1)} + b^{(l)} \right)
\end{align}

\noindent with $K$ being the kernel size, $d$ being the dilation factor, $\sigma$ being nonlinear, and $h_t^{(l)}$ being the hidden activation at layer $l$ and time $t$.
The final classification is performed on the last temporal output, representing the most recent processed information.

%
%%%%%%%%%%%%%%%%%%%%%%%%%
\textbf{Medformer.}
Standard models struggle to capture the multi-scale temporal patterns and variable interactions typical of environmental time series data.
To address this challenge, we include Medformer, a Transformer variant designed for multivariate time series.
Medformer decomposes input into temporal `patches' at multiple resolutions to detect patterns across distinct time scales.
Medformer also learns how different variables influence each other by embedding them together, allowing the model to understand complex relationships across multiple signals. Self-attention mechanisms with 4 heads and 2 layers allow the model to attend to  the most relevant temporal segments.
The attention mechanism uses masking to prevent the model from seeing future time steps:

\begin{align}
Z^l &= \text{softmax}\!\left(\frac{QK^\top}{\sqrt{d_k}} + M\right)V
\end{align}

\noindent where $M$ is a masked attention bias, integrated into the standard transformer attention.
To combine these insights, Medformer applies a two-stage attention mechanism: local attention identifies salient features within each scale while global attention weights contributions of each time scale to the overall representation.
The result is a unified representation that can effectively handle diverse and multi-scale input data.

%
%%%%%%%%%%%%%%%%%%%%%%%%%
\textbf{Spatio-temporal Transformer.}
ST-Transformer extends the Transformer framework by treating each variable as a spatial unit and applying attention jointly across space and time.
Local attention captures interactions among spatial units (e.g., grid cells, stations), while temporal attention tracks dependencies across time.
The input projection layer transforms the original features to higher dimensional vectors suitable for attention mechanisms.
The ST-Transformer first embeds input features, then applies multi-head attention with spatio-temporal encoding:

\begin{align}
Z^0 &= XW_e \\
Q, K, V &= Z^{l-1}W_Q, \; Z^{l-1}W_K, \; Z^{l-1}W_V \\
Z^l &= \text{softmax}\!\left(\frac{QK^\top}{\sqrt{d_k}} + A_{spatial} + A_{temporal}\right)V
\end{align}

\noindent where $l$ is the number of layers, $X$ is input data, $W_e$ is the weight matrix, $A_{spatial}$ encodes positional dependencies, and $A_{temporal}$ encodes temporal dependencies.

The transformer uses 16 attention heads and 3 encoder layers, allowing the model to capture complex temporal relationships across different time scales.
The final classification is performed on the mean-pooled temporal representations, effectively aggregating information across the entire sequence.
This is in contrast to Medformer, which embeds variables jointly to learn cross-signal relationships without assuming spatial organization.
Thus, ST-Transformer is well-suited for hypoxia prediction because it captures both spatial patterns and temporal evolution.
Here, spatial dimensions represent grid cells and temporal dimensions represent observation sequences.

%
%
%%%%%%%%%%%%%%%%%%%%%%%%%%%%%%%%%
%
%
\section{Methodology}
\label{sec:methodology}

%
%%%%%%%%%%%%%%%%%%%%%%%%%
\subsection{Experimental Design and Validation Framework}
\label{sec:design}
We designed a validation framework using multiple models to assess the robustness of deep learning models to predict hypoxic conditions in the Gulf of Mexico (\figref{spatial-domain}).
We partitioned the training and testing data across different time periods, enabling the model to represent temporal characteristics in the data.
Since oxygen dynamics vary by depth, we processed each depth layer independently.
We adopted a multi-model approach by choosing four different neural network architectures that can handle temporal time-series data to help predict hypoxia.
The temporal data are restructured using a sliding window approach that creates sequences of fixed length.
Each prediction target is associated with a preceding sequence of environmental conditions.
We selected a 7-day window size to align with hypoxia events, which typically develop over weekly timescales.
Our training dataset covers the same months as the test ensuring seasonal pattern learning.
The test data are temporarily separated from the training data to provide temporal independence.

% MR To evaluate our hypoxia prediction models, we use several evaluation metrics to assess different aspects of prediction accuracy.
We evaluate classification performance using Receiver Operating Characteristic Area Under the Curve (AUC-ROC), which measures how capable a model is in  distinguishing between normoxic and hypoxic conditions.
Our dataset is highly imbalanced with most instances being non-hypoxic data compared to hypoxic events.
Therefore, we use the Precision Recall Area Under the Curve (AUC-PR), which is particularly well-suited for evaluating model performance on imbalanced datasets.
Finally, we use log loss and the Brier score to understand the quality of predicted probability distribution, crucial for informed decision-making.

Our framework also includes automatic threshold optimization based on maximization of the F1-score.
That is, we compute precision-recall curves across all possible thresholds and identify the optimal decision boundary that balances precision and recall.
This optimization is performed independently for each model, ensuring that each architecture operates at its optimal decision threshold.
We show the confusion matrices for the optimized thresholds, providing insight into the overall classification performance.
The implementation computes true positives, false positives, true negatives, and false negatives, enabling detailed error analysis.
This analysis is particularly important for prediction of hypoxic events, where false negatives (missed hypoxia events) and false positives (false alarms) could lead to misinformed decisions or other serious consequences.

%
%%%%%%%%%%%%%%%%%%%%%%%%%
\subsection{Implementation Details}
\label{sec:implementation}
The custom data loader implementation incorporates weighted random sampling to maintain class balance during training.
This complements SMOTE oversampling (\sectref{pipeline}), which addresses dataset-level imbalance while weighted sampling ensures batch-level balance.
The sample weights are computed on the basis of the class frequencies, ensuring that each batch contains representative samples from both classes.
To achieve class balance while maintaining the temporal integrity of the data sequences, we use PyTorch's \texttt{WeightedRandomSampler}.
%
% \magesh{Insert something here about the ST-Transformer: grids are spatial ($A_{spatial}$) and temporal corresponds to 7-day sequences ($A_{temporal}$). We need to communicate the spaital and temporal dimensions for Reviewer 1, Comment 1}
%
ST-Transformer encodes spatial relationships between grid cells using spatial ($A_{spatial}$), and temporal relationships along 7-day sequences using ($A_{temporal}$), allowing simulational learning of where hypoxia occurs and how it evolves over time.

To ensure stable convergence when learning complex temporal patterns in oceanographic data, the training implementation uses the Adam optimizer, the learning rate being set to 0.001.
The learning rate is kept constant throughout the training to maintain consistent gradient updates, which is useful for modeling temporal relationships in the data.
We do not use batch normalization since it can disrupt temporal dependencies in sequential data.

The training implementation processes data in batches of 1024 sequences, balancing computational efficiency with memory constraints.
Each epoch processes the entire training dataset, with loss computation performed on each batch.
The implementation includes comprehensive log of training progress, including per-epoch loss values and convergence monitoring.
\textit{Early Stopping} is not implemented, as the temporal nature of the data requires full training to capture all seasonal patterns over multiple annual cycles.
Our temporally separated test sets (August 2020, Summers 2022-2024) rigorously evaluate generalization without validation-based stopping.
We want to ensure that the learned models can be used to simulate real-world deployments where models can predict future conditions.
We successfully validated real-world deployment by testing the model with real-world data from 2022-2024. 
% The source code is available at \cite{code}.

%
%
%%%%%%%%%%%%%%%%%%%%%%%%%%%%%%%%%
%
%
\section{Experimental Results and Discussion}
\label{sec:experimental}
We trained the models for different numbers of epochs, although epoch 30 was the best for each model.
% We do not perform \textit{EarlyStopping} since it is necessary for our models to learn seasonal progression over multiple annual cycles.
We then tested the trained model with the test set for August 2020 as well as summer data for 2022 to 2024.

%
%%%%%%%%%%%%%%%%%%%%%%%%%
\subsection{Classification Performance Analysis}

\begin{table}[t!]
\centering
\caption{Hypoxia classification results for 4 timeframes.
}
\label{tab:performance_datasets}
\resizebox{\columnwidth}{!}{%
\begin{tabular}{llcccc}
\toprule
\textbf{Dataset} & \textbf{Model} & \textbf{AUC-ROC} & \textbf{AUC-PR} & \textbf{Accuracy} & \textbf{F1} \\
\midrule
\multirow{4}{*}{\begin{tabular}[c]{@{}l@{}}August 2020\end{tabular}} 
& BiLSTM         & 0.986 & 0.825 & \textbf{0.974} & 0.744 \\
& Medformer      & 0.976 & 0.806 & 0.972 & 0.738 \\
& ST-Transformer & \textbf{0.992} & \textbf{0.881} & 0.948 & \textbf{0.802} \\
& TCN            & 0.974 & 0.784 & 0.970 & 0.712 \\
\midrule
\multirow{4}{*}{\begin{tabular}[c]{@{}l@{}}Summer 2022\end{tabular}}
& BiLSTM         & 0.972 & 0.746 & 0.925 & 0.716 \\
& Medformer      & 0.955 & 0.755 & 0.656 & 0.702 \\
& ST-Transformer & \textbf{0.983} & \textbf{0.831} & \textbf{0.943} & \textbf{0.774} \\
& TCN            & 0.947 & 0.694 & 0.812 & 0.658 \\
\midrule
\multirow{4}{*}{\begin{tabular}[c]{@{}l@{}}Summer 2023\end{tabular}}
& BiLSTM         & 0.981 & 0.736 & 0.935 & 0.704 \\
& Medformer      & 0.944 & 0.702 & 0.564 & 0.670 \\
& ST-Transformer & \textbf{0.988} & \textbf{0.820} & \textbf{0.957} & \textbf{0.781} \\
& TCN            & 0.956 & 0.720 & 0.614 & 0.701 \\
\midrule
\multirow{4}{*}{\begin{tabular}[c]{@{}l@{}}Summer 2024\end{tabular}}
& BiLSTM         & 0.976 & 0.730 & 0.949 & 0.709 \\
& Medformer      & 0.962 & 0.718 & 0.877 & 0.661 \\
& ST-Transformer & \textbf{0.982} & \textbf{0.782} & \textbf{0.959} & \textbf{0.741} \\
& TCN            & 0.967 & 0.674 & 0.936 & 0.680 \\
\bottomrule
\end{tabular}%
}
\end{table}

We evaluated model performance on the basis of a number of metrics.
We develop and test a binary classification model that predicts hypoxia or non-hypoxia.
The core performance metrics of a model is calculating the accuracy of the model itself, i.e. given the unseen data how well the model is able to predict true values.
Model accuracy represents the fraction of correct predictions, although this metric can be misleading with imbalanced datasets.
Table \ref{tab:performance_datasets} shows model accuracy for each dataset.
BiLSTM achieved the highest accuracy (97.4\%) for August 2020 test data, but accuracy alone can be misleading for imbalanced datasets.
%
% \begin{equation}
%     Accuracy = \frac{TP + TN}{TP + TN + FP + FN}
% \end{equation}

% where TP - True Positive, TN - True Negative, FP - False Positive, FN - False Negative.
%

\begin{figure}[t!]
\centering
\includegraphics[width=0.95\columnwidth]{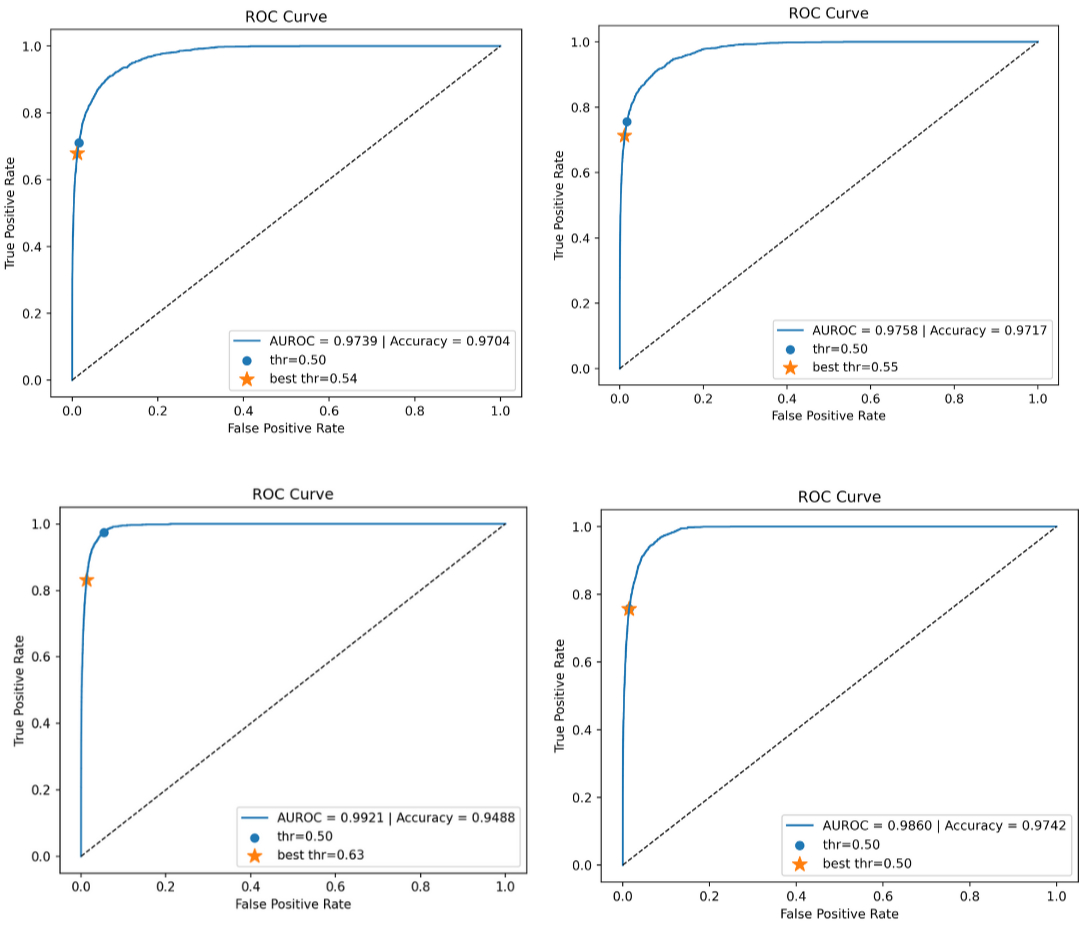}
\caption{ROC Curve of all 4 models. \textit{TCN} (Top Left), \textit{Medformer} (Top Right), \textit{STT} (Bottom Left), \textit{BiLSTM} (Bottom Right)}
\label{fig:Models-ROC}
\end{figure}

\textbf{AUC-ROC.}
Since hypoxic events are rare, high accuracy may reflect correct prediction of abundant normoxic conditions rather than effectively detecting hypoxic ones.
% Therefore, we use ROC analysis to better assess model performance.
% Since we are working with an imbalanced dataset, it is important to understand the measure of positive and negative instances across classes.
An ROC curve shows the variation of  the True Positive Rate with respect to the False Positive Rate for different values of the threshold, and the AUC represents the area under this curve, measures how effectively the  model prioritizes  positive instances over negative ones.
% MR This metric calculates a model's ability to distinguish between classes across all classification thresholds. 

Figure \ref{fig:Models-ROC} shows different ROC curve thresholds for each model using August 2020 data.
We observe in Figure \ref{fig:Models-ROC} that all models show strong discriminative ability with AUROC values ranging from 0.9739 to 0.9921 while BiLSTM is the most balanced with high AUROC (0.9860) and accuracy (0.9742).

% \begin{equation}
%     Precision = \frac{TP}{TP + FP}
% \end{equation}

% \begin{equation}
%     Recall = \frac{TP}{TP + FN}
% \end{equation}

% \begin{equation}
%     F1 Score = 2 \times \frac{Precision \times Recall}{Precision + Recall}
% \end{equation}

\textbf{Precision-recall and F1.}
AUC-ROC can be overly optimistic with class imbalance, so we also use precision-recall for model evaluation.
Precision reflects the fraction of  all predicted positives that were correctly identified, while recall indicates how effectively all true positives are detected.
The F1 score is the harmonic mean of precision and recall, providing a single measure that balances how many predicted positives are correct with how many actual positives are recovered. 
Lower precision (recall) values suggest a stronger propensity to overpredict (underpredict) hypoxic conditions.

\begin{figure}[t!]
\centering
\includegraphics[width=0.99\columnwidth]{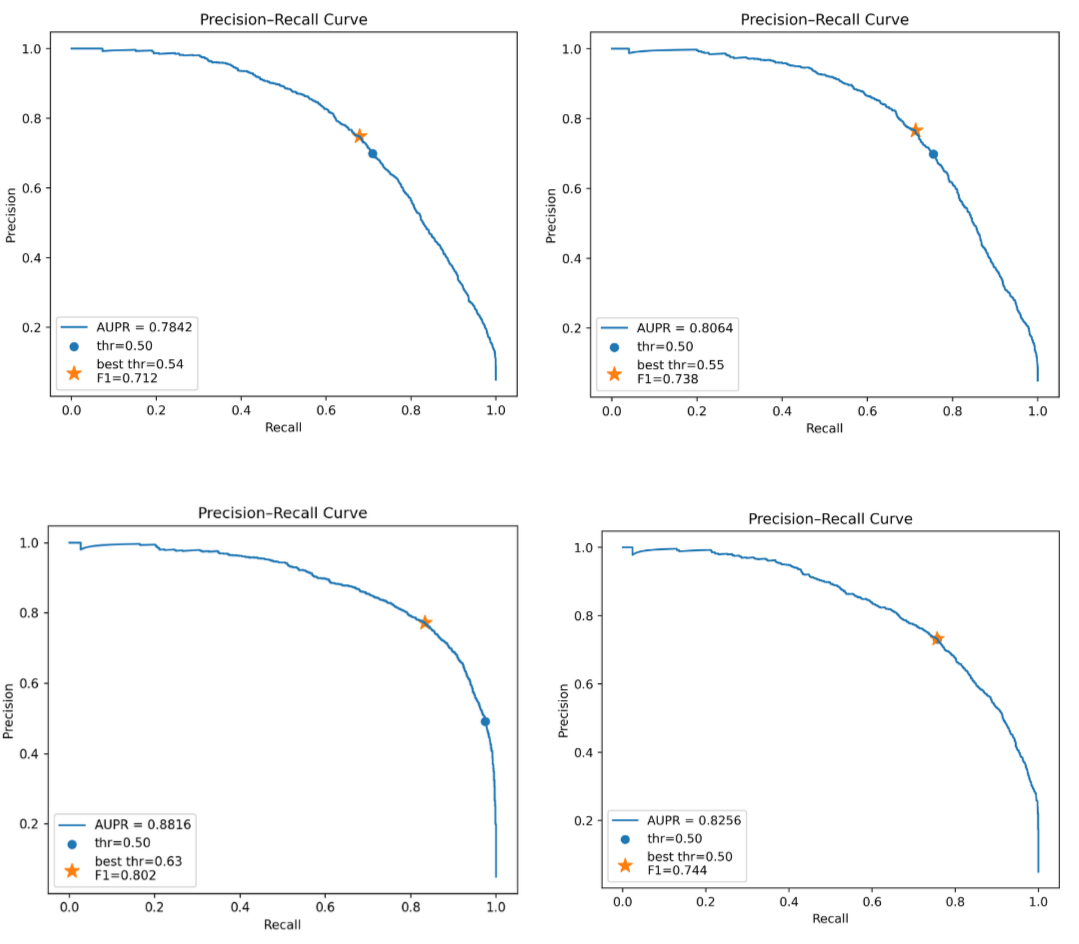}
\caption{PR Curve of all 4 models. \textit{TCN} (Top Left), \textit{Medformer} (Top Right), \textit{STT} (Bottom Left), \textit{BiLSTM} (Bottom Right)}
\label{fig:Models-PR}
\end{figure}

\begin{figure}[t!]
\centering
\includegraphics[trim=10 0 0 0, clip, width=0.95\columnwidth]{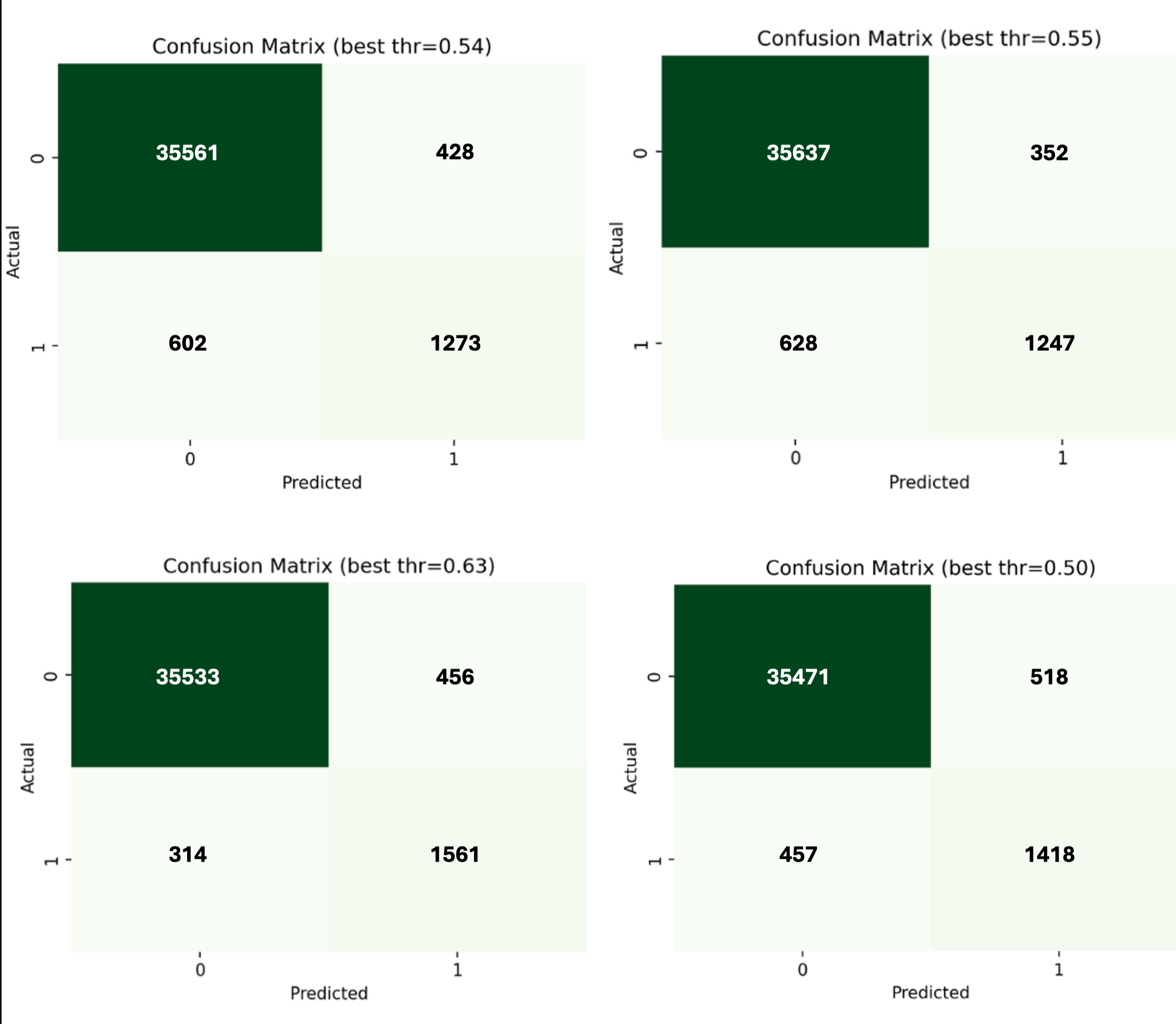}
\caption{Confusion matrices of all 4 models at optimized thresholds. \textit{TCN} (Top Left), \textit{Medformer} (Top Right), \textit{ST-Transformer} (Bottom Left), \textit{BiLSTM} (Bottom Right)}
\label{fig:CM}
\end{figure}

\textbf{Confusion matrices.}
Figure \ref{fig:Models-PR} shows the precision-recall curves for each model at different thresholds; '*' indicates the best threshold in both Figures \ref{fig:Models-PR} and \ref{fig:Models-ROC}.
We also present confusion matrices (Figure~\ref{fig:CM}) at their optimized thresholds.

Identifying hypoxia in real-world operational settings is challenging.
Thus, we are most interested in correct prediction of true negatives to avoid false alarms (predicting hypoxia when conditions are normoxic).
ST-Transformer predicted the largest number of True Negatives ($1561$) at optimized threshold versus $1320$ at the default threshold.
This demonstrates that the ST-Transformer model, when combined with the spatial and temporal signals, better learns the pattern of hypoxia and avoids false alarams.

\textbf{Brier Score.}
We also use calibration metrics to assess model prediction quality.
\textit{Brier Score} (BS) provides the mean-squared error of probabilistic predictions for binary outcomes.
A lower Brier score indicates more accurate predictions ranging from $0.0$ (perfect) to $1.0$ (worst possible):

\begin{equation}
BS = \frac{1}{N} \sum_{i=1}^{N} (p_i - y_i)^2
\end{equation}

\noindent with $N$ being the cardinality of the set of samples, $p_i \in [0,1]$ being the predicted probability of the positive class, 
and $y_i \in \{0,1\}$ being the ground truth outcome.

\begin{table}[t!]
\centering
\caption{Log Loss and Brier Score for 4 timeframes across each model's prediction. 
}
\label{tab:model_performance}
\resizebox{\columnwidth}{!}{%
\begin{tabular}{llccc}
\toprule
\textbf{Dataset} & \textbf{Model} & \textbf{Brier Score} & \textbf{Log Loss} \\
\midrule
\multirow{4}{*}{\begin{tabular}[c]{@{}l@{}}August 2020\end{tabular}} 
& \textbf{BiLSTM}        & \textbf{0.0215} & \textbf{0.0846} & \\
& Medformer      & 0.0291 & 0.1252 &  \\
& ST-Transformer & 0.0282 & 0.1422 & \\
& TCN            & 0.0237 & 0.0904 & \\
\midrule
\multirow{4}{*}{\begin{tabular}[c]{@{}l@{}}Summer 2022\end{tabular}}
& BiLSTM         & 0.0589 & 0.2117 &  \\
& Medformer      & 0.2043 & 0.5902 &  \\
& ST-Transformer & 0.0686 & 0.2663 &  \\
& TCN            & 0.1386 & 0.4443 &  \\
\midrule
\multirow{4}{*}{\begin{tabular}[c]{@{}l@{}}Summer 2023\end{tabular}}
& BiLSTM         & 0.0532 & 0.1953 &  \\
& Medformer      & 0.2417 & 0.6733 &  \\
& ST-Transformer & 0.0580 & 0.2367 &  \\
& TCN            & 0.2612 & 0.7812 &  \\
\midrule
\multirow{4}{*}{\begin{tabular}[c]{@{}l@{}}Summer 2024\end{tabular}}
& BiLSTM         & 0.0416 & 0.1591 &  \\
& Medformer      & 0.0915 & 0.3069 &  \\
& ST-Transformer & 0.0530 & 0.2219 &  \\
& TCN            & 0.0575 & 0.2147 & \\
\bottomrule
\end{tabular}%
}
\end{table}

We observe the BS value of each model across our 4 test datasets in Table \ref{tab:model_performance}.
For August 2020 test data, all models showed low BS indicating good performance.
As an example of high variability due to real-world test data, consider Medformer.
With August 2020 data we have BS $0.0291$ compared to the real-world data (Summers 2022-24) in which Medformer achieved the highest BS of all models ($0.2417$)
In constrast, BiLSTM and ST-Transformer were consistent across all test periods with BS varying by $\pm  0.037$ and $\pm 0.040$, respectively.
This demonstrates the reliability of BiLSTM and ST-Transformer for operational deployment.

\textbf{Log loss.}
Log loss (cross-entropy) is a calibration metric that evaluates probabilistic prediction accuracy by penalizing confident wrong predictions more heavily:

\begin{equation}
\text{LogLoss} = -\frac{1}{N} \sum_{i=1}^{N} \left[ z_i \log(q_i) + (1-z_i)\log(1-q_i) \right]
\end{equation}

\noindent where $N$ is the number of samples, where $q_i$ is the probability predicted  for the $i$th sample, and $z_i$ is the observed binary label.

Like BS, log loss ranges from $0.0$ (perfect predictions) to infinity (confident but incorrect predictions).
For a balanced binary result, random guessing with $p=0.5$ yields a log loss of approximately $0.693$. 

In Table \ref{tab:model_performance}, a log loss value greater than 0.693 was only achieved by TCN ($0.7812$) with Summer 2023 data, although Medformer was close ($0.6733$).
This indicates that TCN peformed worse than random guessing on Summer 2023 data.
In contrast, low log loss across all tests for BiLSTM and ST-Transformer indicates strong performance with real-time data. 

%
%%%%%%%%%%%%%%%%%%%%%%%%%
\subsection{Pairwise Model Comparison with McNemar's Test}

We apply McNemar's test to determine whether the performance differences between models are statistically significant.
McNemar's test is appropriate for binary classification comparisons. 
The implementation constructs contingency tables capturing patterns of agreement and disagreement between model predictions.
The test statistic is calculated as:

\begin{equation}
\chi^2 = \frac{(b - c)^2}{b + c}
\end{equation}

\noindent with $b$ and $c$ respectively being the number of discordant pairs having outcomes $(1,0)$ and $(0,1)$.
$p$-values are computed by using the cumulative distribution function of the $\chi^2$ distribution for each pair of models.

\begin{figure}[t!]
\centering
\includegraphics[trim=0 0 10 0, clip, width=0.99\columnwidth]{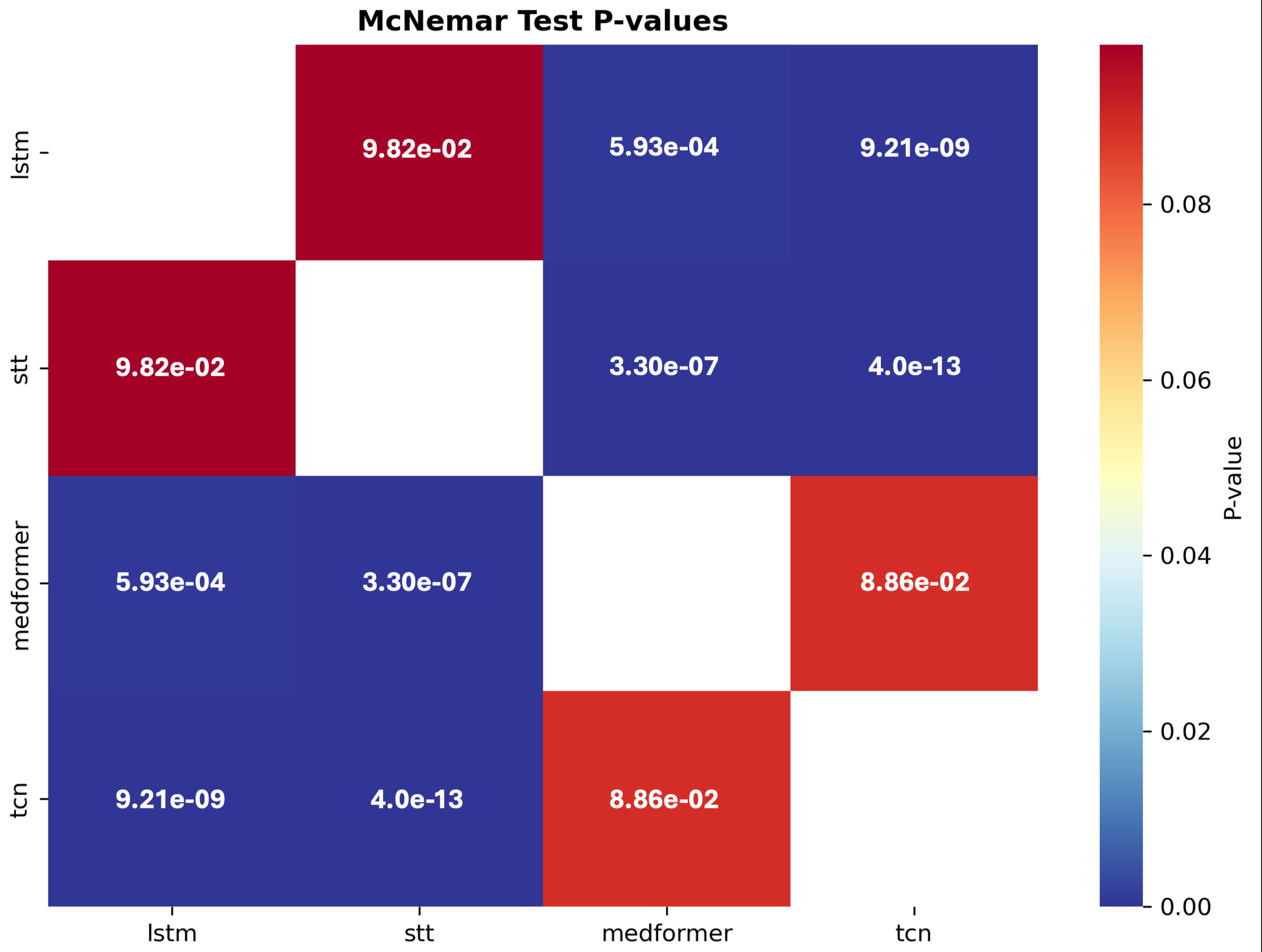}
\caption{Pairwise McNemar test $p$-values across all models using August 2020 test data.}
\label{fig:Models-sig}
\end{figure}

Figure \ref{fig:Models-sig} shows the $p$-values for all pairwise comparisons using August 2020 test data.
Four out of six model pairs demonstrated particularly strong significance ($p < 0.001$) observed for several comparisons including ST-Transformer vs TCN ($p=4.0e{-13}$), ST-Transformer vs Medformer ($p=3.30e{-07}$), BiLSTM vs TCN ($p=9.21e{-09}$), and BiLSTM vs Medformer ($p=5.93e{-04}$).
However, differences between BiLSTM and ST-Transformer ($p=0.0982$) and between Medformer and TCN ($p=0.0886$) did not reach the conventional $0.05$ significance threshold, indicating these model pairs have statistically similar predictions.

% This indicates statistically significant differences between all models when tested on the August 2020 data.

\begin{table}[t!]
\centering
\caption{Pairwise effect sizes (Cohen's $w$) between models ($p<0.001$).
}
\label{table:effect-size}
\resizebox{\columnwidth}{!}{%
\begin{tabular}{llcccc}
\toprule
\textbf{Dataset} & \textbf{Model} &  &  &  &  \\
 &  & BiLSTM & Medformer & ST-Transformer & TCN \\
\midrule
\multirow{4}{*}{\begin{tabular}[c]{@{}l@{}}Summer 2022\end{tabular}}
& BiLSTM         & - & 0.9962 & \textbf{1.5288} & 1.2687 \\
& Medformer      & 0.9962 & - & 1.0243 & 0.9046 \\
& ST-Transformer & 1.5288 & 1.0243 & - & 1.2890 \\
& TCN            & 1.2687 & 0.9046 & 1.2890 & - \\
\midrule
\multirow{4}{*}{\begin{tabular}[c]{@{}l@{}}Summer 2023\end{tabular}}
& BiLSTM         & - & 0.9101 & \textbf{1.5655} & 0.9582 \\
& Medformer      & 0.9101 & - & 0.9448 & 0.6467 \\
& ST-Transformer & 1.5655 & 0.9448 & - & 0.9875 \\
& TCN            & 0.9582 & 0.6467 & 0.9875 & - \\
\midrule
\multirow{4}{*}{\begin{tabular}[c]{@{}l@{}}Summer 2024\end{tabular}}
& BiLSTM         & - & 1.4204 & \textbf{1.5922} & 1.5547 \\
& Medformer      & 1.4204 & - & 1.4333 & 1.4236 \\
& ST-Transformer & 1.5922 & 1.4333 & - & 1.5626 \\
& TCN            & 1.5547 & 1.4236 & 1.5626 & - \\
\bottomrule
\end{tabular}%
}
\end{table}

\textbf{Effect size analysis.}
While McNemar's test revealed statistically significant differences between model pairs, effect size measures the magnitude of those differences.
Effect size helps identify which model strikes the most favorable balance  between accuracy and computational overhead, even in cases where statistical significance was not achieved.

In \tableref{effect-size} we report effect sizes between models derived from McNemar's test with $p < 0.001$.
We use Cohen's $w$ as the measure of effect size:

\begin{equation}
w = \sqrt{\frac{\chi^2}{N}}
\end{equation}

\noindent where $\chi^2$ is the chi-square statistic from McNemar's test 
and $N$ is the total sample size.
We observe that all pairwise comparison values exceeded Cohen's threshold for large effects (commonly $\geq 0.5$).
This indicates that the performance difference between models cannot be attributed to minor variations.

In terms of model-specific patterns, the BiLSTM and ST-Transformer comparison resulted in the highest Cohen's $w$ values in each year (1.5288, 1.5655, and 1.5922, respectively for Summer 2022 through 2024).
The high values indicate that BiLSTM and ST-Transformer fundamentally capture different aspects of the underlying temporal patterns.
In comparison, the Medformer and TCN Cohen's $w$ values showed the smallest effect sizes, but they are still substantial ranging from 0.6467 to 1.4236.

\begin{figure}[t!]
\centering
\includegraphics[trim=10 50 10 10, clip, width=0.99\columnwidth]{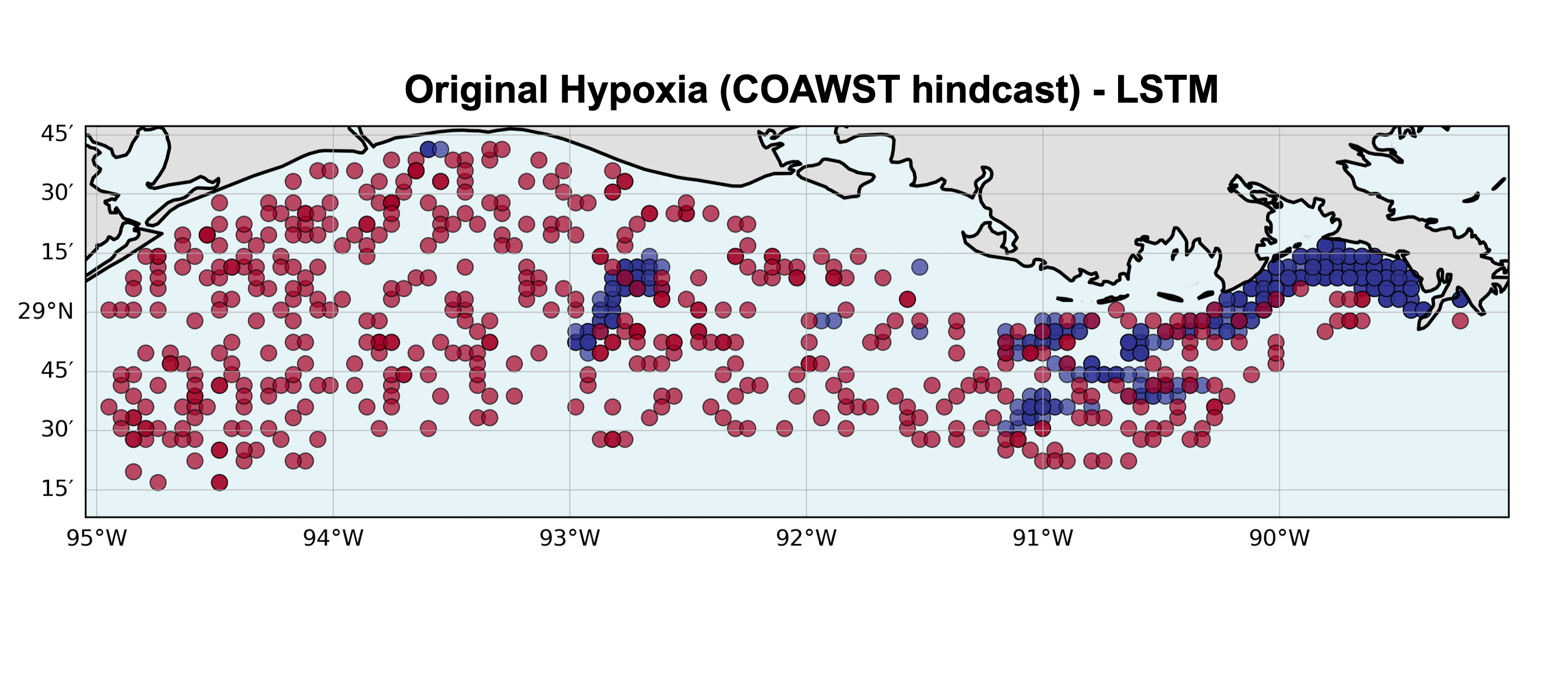}
\caption{ROMS plot of August 2020 test data: COAWST hindcast showing hypoxia (blue dots) and normoxia (red dots).}
\label{fig:observed-aug2020}
\end{figure}

\begin{figure}[t!]
\centering
\includegraphics[trim=0 250 0 150, clip, width=0.99\columnwidth]{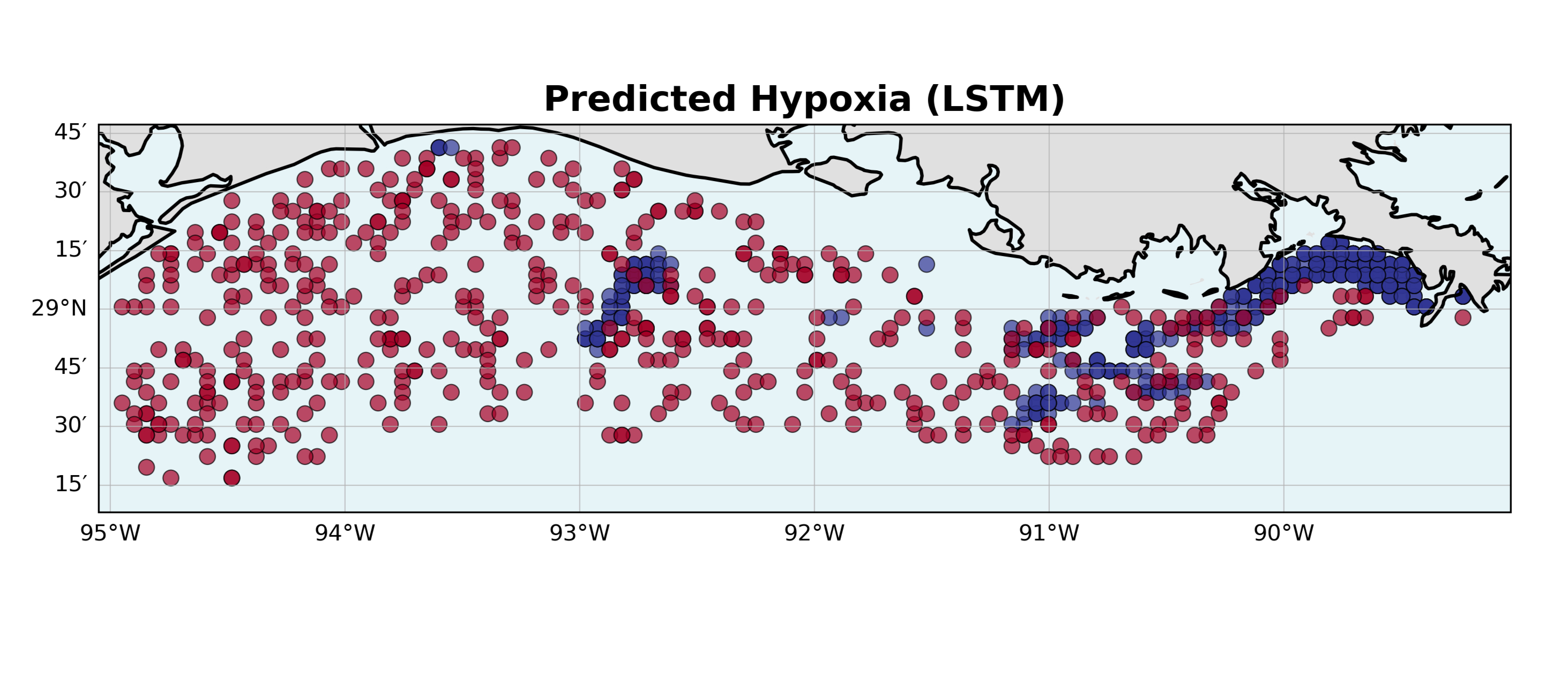}
\caption{ROMS plot of August 2020 test data predicted using BiLSTM.}
\label{fig:lstm-predicted-aug2020}
\end{figure}

\begin{figure}[t!]
\centering
\includegraphics[trim=0 250 0 150, clip, width=0.99\columnwidth]{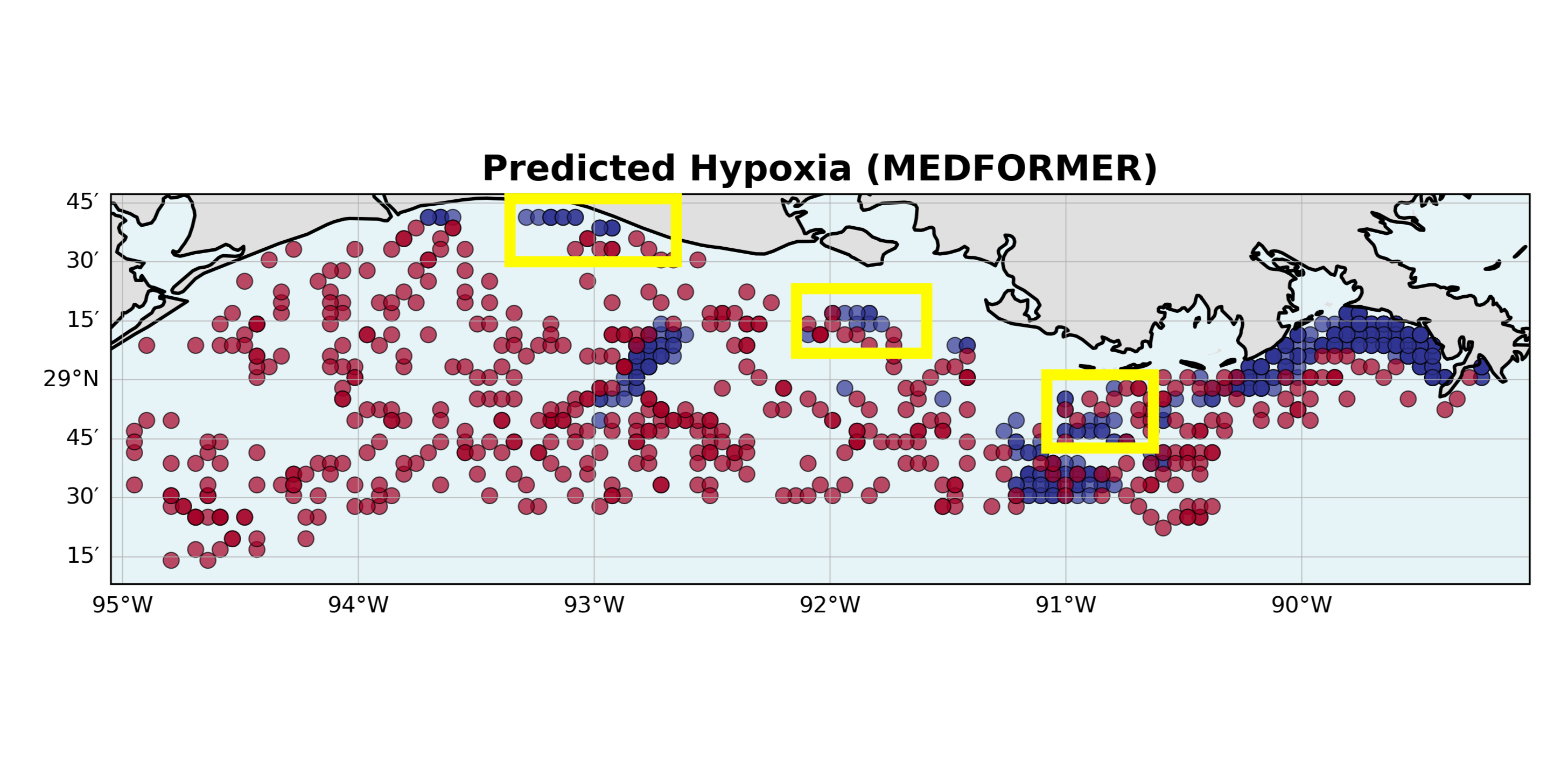}
\caption{ROMS plot of August 2020 test data predicted using MedFormer.}
\label{fig:med-predicted-aug2020}
\end{figure}

\subsection{Spatial Validation}
Spatial validation determines whether a model can capture the underlying oceanographic processes.

We validated model predictions by plotting them against ROMS hindcast spatial data.
%The number of samples here are populated at random.
Figure \ref{fig:observed-aug2020} shows a random sample of COAWST hindcast hypoxia (blue) and normoxia (red) for August 2020 test data.
We see a group of blue dots indicating hypoxic conditions (oxygen $< 2.0mg/L$) concentrated near the Gulf shores.
This coastal pattern reflects nutrient loading and Mississippi River freshwater outflow.

Figures \ref{fig:lstm-predicted-aug2020} and \ref{fig:med-predicted-aug2020} show BiLSTM and Medformer predictions, respectively, for the same test data.
BiLSTM predictions align well with the events observed in Figure \ref{fig:observed-aug2020}.
However, Medformer shows clear misclassifications highlighted with yellow boxes in Figure \ref{fig:med-predicted-aug2020}.
Consistent with Table \ref{tab:performance_datasets}, BiLSTM is spatially superior to Medformer with the test data.

\begin{figure}[t!]
\centering
\includegraphics[trim=10 10 5 10, clip,width=0.99\columnwidth]{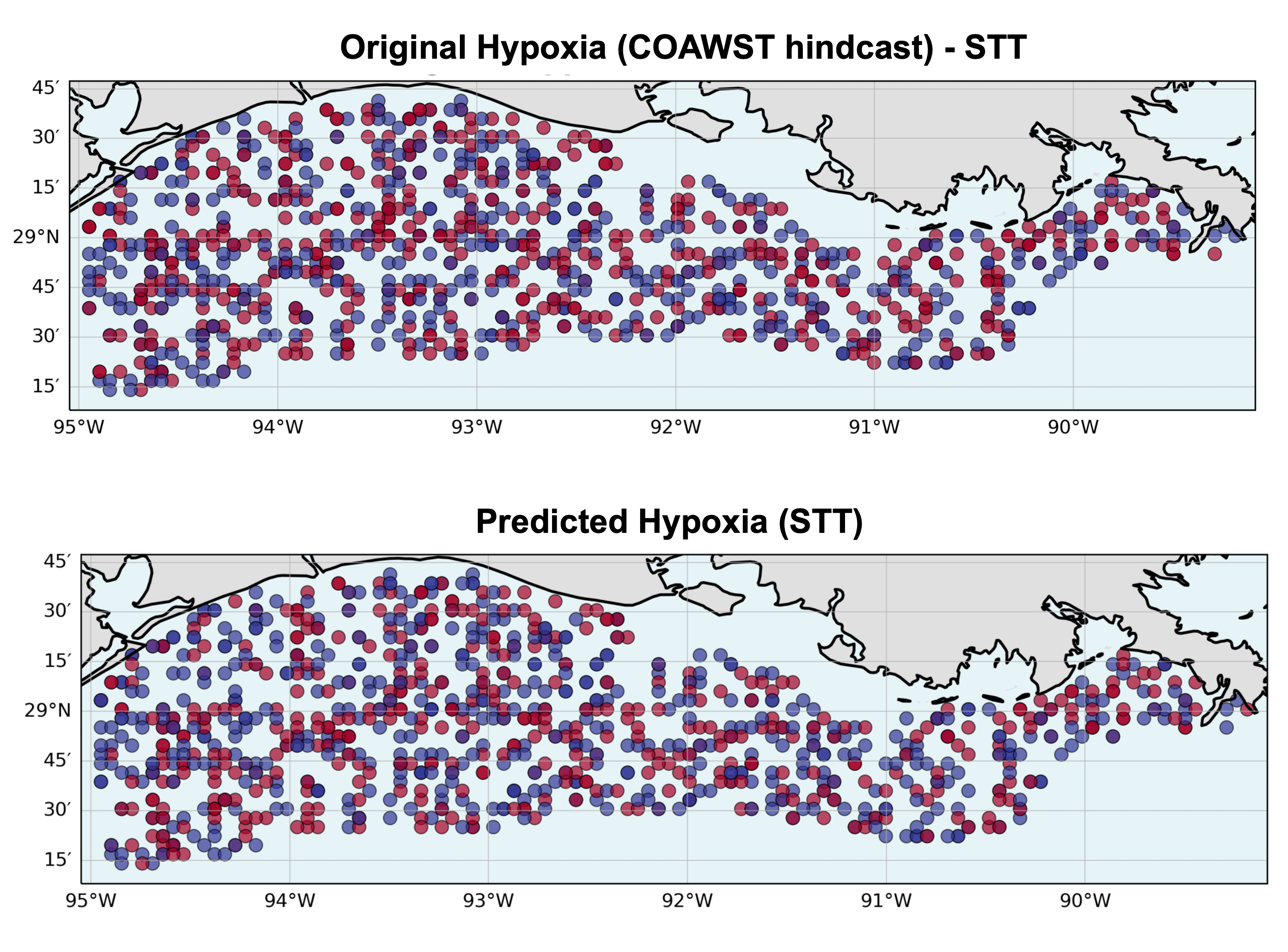}
\caption{ROMS hindcast validation: COAWST hindcast versus ST-Transformer (best model) predictions for Summer 2022 (1000 random points). }
\label{fig:roms-stt-summer-2022}
\end{figure}

\begin{figure}[t!]
\centering
\includegraphics[trim=10 10 10 10, clip, width=0.99\columnwidth]{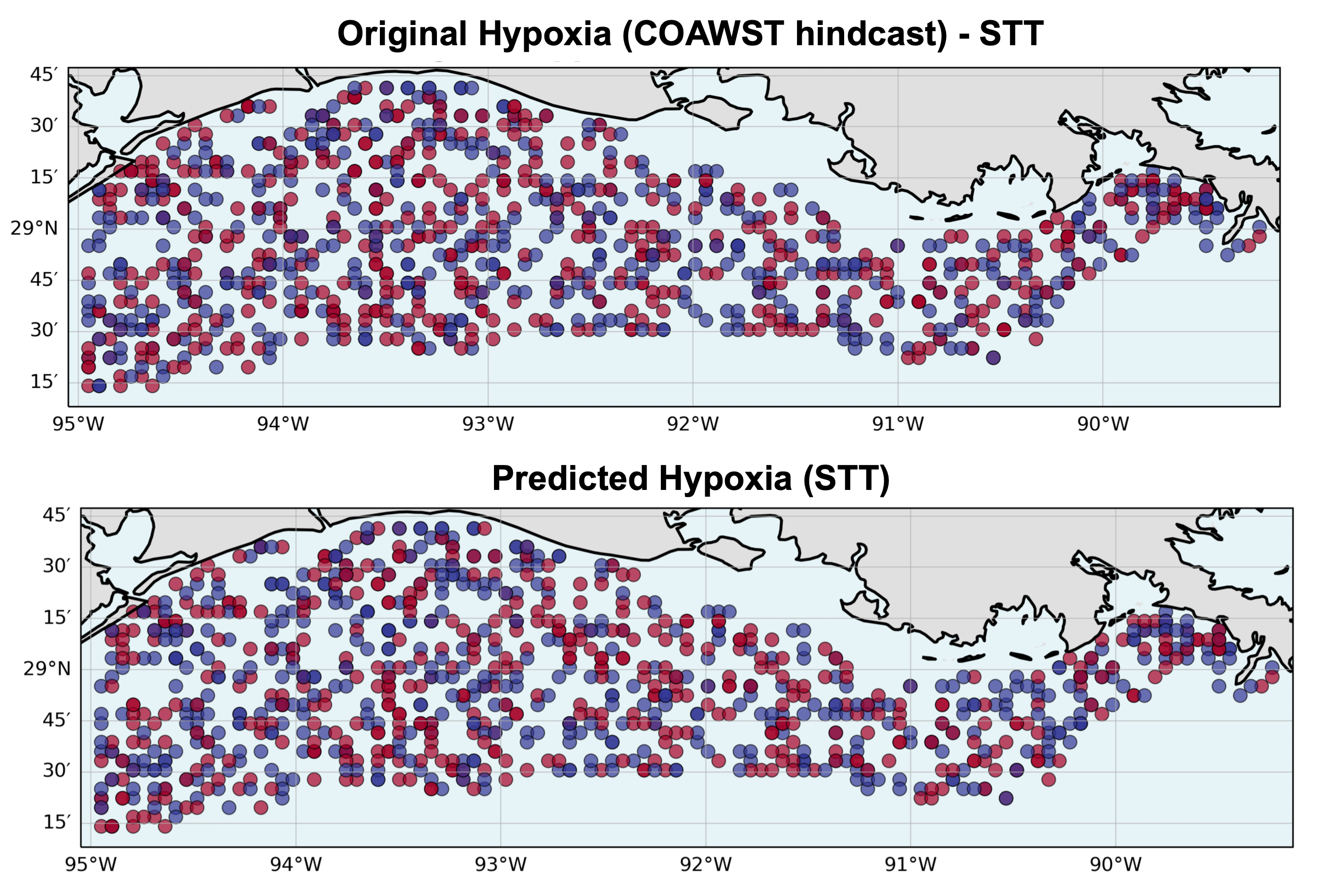}
\caption{ROMS hindcast validation: COAWST hindcast versus ST-Transformer (best model) predictions for Summer 2023 (1000 random points).}
\label{fig:roms-stt-summer-2023}
\end{figure}

\begin{figure}[t!]
\centering
\includegraphics[trim=50 130 25 50, clip, width=0.99\columnwidth]{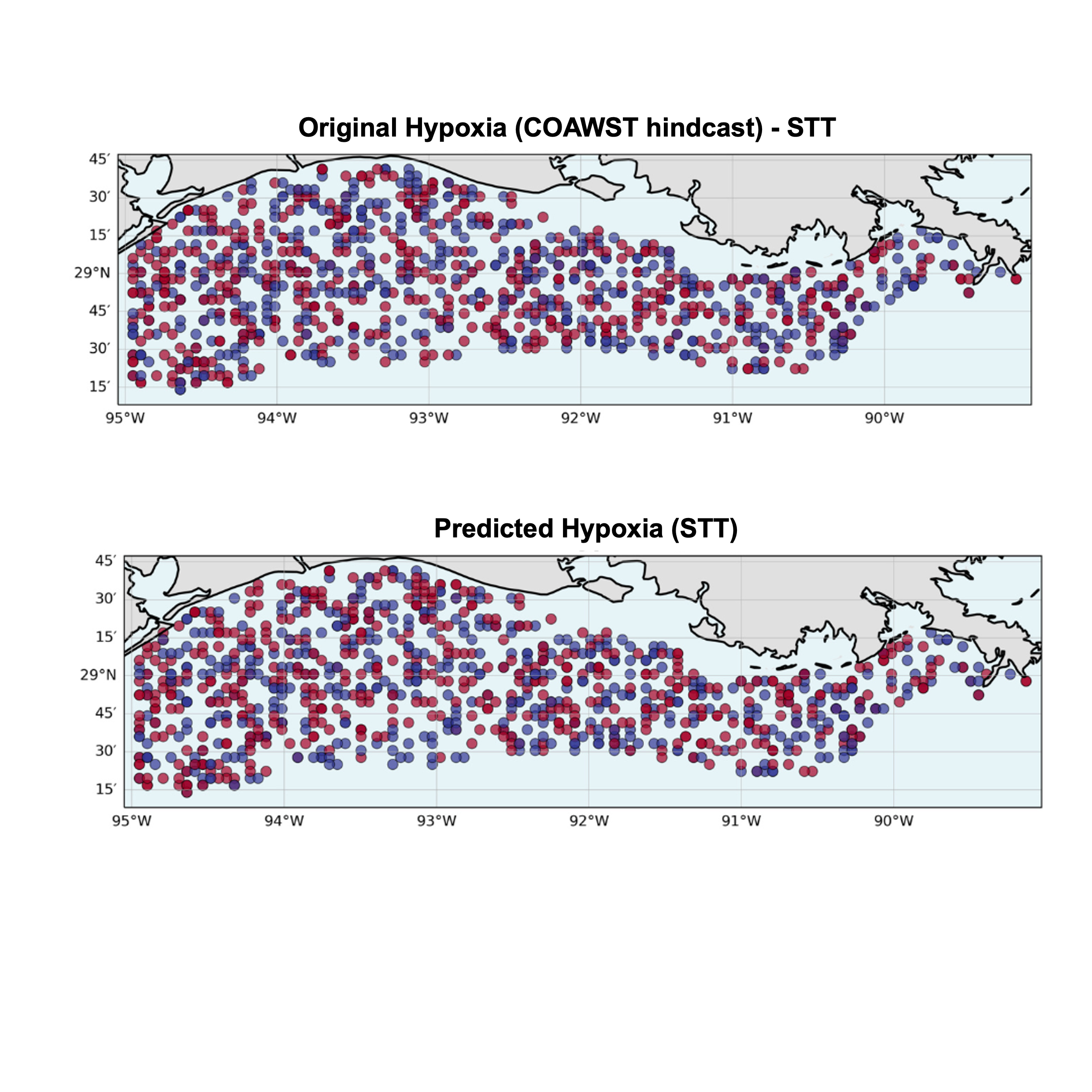}
\caption{ROMS hindcast validation: COAWST hindcast versus ST-Transformer (best model) predictions for Summer 2024 (1000 random points).}
\label{fig:roms-stt-summer-2024}
\end{figure}

ST-Transformer demonstrates consistent spatial accuracy across our 2022-2024 test data; see Figures \ref{fig:roms-stt-summer-2022}-\ref{fig:roms-stt-summer-2024}.
ST-Transformer accurately captures coastal hypoxia patterns attributed to river-plume dynamics and stratification.
These results support operational deployment of ST-Transformer and BiLSTM for fishery management and field research planning.

%
%
%%%%%%%%%%%%%%%%%%%%%%%%%%%%%%%%%
%
%
\section{Related Works}
\label{sec:related}
Different types of AI models have been used for diverse applications ranging from geospatial to robotics \cite{sharma2018complete,basu2015deepsat,Ou2025}. Below, we survey different AI-based approaches for coastal hypoxia and related applications. 
\textbf{Hypoxia prediction approaches.}
Hypoxia prediction has evolved from seasonal statistical models to daily-resolution systems for the Gulf of Mexico.
Katin et al. \cite{katin2022} introduced daily Bayesian mechanistic forecasting explaining about 50\% of the variability in hypoxic area.
% They rigorously reported measures of uncertainty and did not report any accuracy metrics.
Ou et al. \cite{ou-hydrodynamic} developed ensemble regression ($R^2 = 0.92$) combining generalized linear models (GLMs) and generalized additive models (GAMs).
Using satellite-derived variables with a Random Forest model, Li et al.~\cite{LI2023113346} predicted bottom dissolved oxygen with $\pm1.2-1.4$~mg/L accuracy.
% They also identified time lags of approximately 0--5 and 16--19 days linking surface processes to bottom hypoxia.

Ou et al. \cite{Ou2025} proposed a multi-classifier model combining U-net and DeepLabv3+ for Louisiana-Texas shelf spatial hypoxia prediction.
Their approach achieved remarkable computational speedup from 1800s to 1s all while maintaining ROMS-compatible accuracy using the same environmental variables (PEA, sediment oxygen consumption, and temperature-dependent decomposition).
However, their spatial segmentation approach treats daily forecasts independently, thus missing the temporal dependencies that result in hypoxia development over  multiple days.
This work is most similar to ours, but our sequence-based approach was developed to capture these temporal patterns driving hypoxic events.

\textbf{Time series deep learning.}
In the literature, architecture comparison studies reveal performance dependencies on temporal characteristics.
P\"{o}lz et al.~\cite{polz2024} demonstrated that a Transformer outperformed a LSTM by about 9\% for longer-timescale hydrological systems but underperformed by roughly 4\% for shorter timescales.
At six locations near China, Fu et al.~\cite{fu2024prediction} demonstrated that a hybrid LSTM-Transformer achieved superior performance compared to individual architectures for prediction of sea surface temperature.
Yan et al.~\cite{yan2020temporal} reported that TCNs achieved competitive skill for El Ni\~no-Southern Oscillation (ENSO) forecasting across multiple lead times (1 month, 3 months, 6 months, and 12 months \cite{yan2020temporal}), demonstrating the ability of TCNs to effectively capture long-range dependencies for climate prediction tasks.
Wu et al. \cite{wu2024} introduced physics-informed Transformers for ocean temperature and salinity prediction, while Zhang et al. \cite{Zhang2025} developed spatio-temporal attention mechanisms applied to ambient air pollution forecasting.
Together, these and other works demonstrate that Transformer-based architectures are well-suited for environmental applications. 

% \textbf{Environmental big data analytics.}
% Prediction of environmental factors increasingly depends on the integration of remote sensing capabilities (i.e., satellite) and in-situ observational networks to capture real-time, high-resolution data across diverse ecosystems.
% The VISION framework exemplifies this approach, offering a virtual integration platform that simulates in-situ observations to complement satellite data and enhance model fidelity across scales \cite{russo2025vision}.
% Real-time coastal monitoring systems require integration of satellite imagery, sensor networks, and numerical models into a unified pipeline.
% However, these systems must operate under strict latency constraints and thus are expensive for operational deployment.

% We did not perform efficiency analyses or report such results.

% Despite the growing complexity and scale of workflows around environmental data, comparative evaluations of computational efficiency remains limited.
% Benchmarking these architectures under environmental workloads is still sparse, despite the increasing use of deep-learning models for predicting climate-related phenomena.
% A recent study by Alotaibi and Nassif underscores the need for explainable and efficient AI models in environmental monitoring, emphasizing challenges like model transparency and data quality in resource-constrained regions \cite{Alotaibi2024}.

\textbf{Research gaps and our contributions.}
A comparison of temporal deep learning architectures for Gulf hypoxia prediction under identical conditions is lacking. We also observe other gaps in the literature. Medformer and spatio-temporal attention models (i.e., Transformer model variants) remain unevaluated for coastal hypoxia prediction.
In addition, environmental AI studies do not often compare performance of architectures using formal statistical significance tests.
To address both of these issues, we applied McNemar’s test (\sectref{experimental}) on paired classification outcomes to evaluate the differences observed across four different architectures, including two Transformer architectures.
% MR In this work, we benchmark four temporal architectures using a Gulf hypoxia dataset with identical preprocessing, splits, and metrics.

% \textbf{G4:} Despite its importance for near–real-time operations, computational efficiency reporting (training/inference time, memory, etc.) is limited.

%\textbf{Our contributions.}

% \mr{Last, to address \textbf{G4}, we report computational efficiency for training and inference (wall-clock time and memory footprint), along with hardware and implementation details to enable evidence-based selection of models for operational deployments.}

%
%
%%%%%%%%%%%%%%%%%%%%%%%%%%%%%%%%%
%
%
\section{Conclusions}
\label{sec:conclusions}

We compared four deep learning architectures for daily coastal hypoxia forecasting using consistent data preparation, experimental conditions, and evaluation protocols.
We found ST-Transformer achieved the highest AUC-ROC values (0.982-0.992) and the highest metrics across all test periods (2020, 2022-2024).
While McNemar's test indicated that differences between ST-Transformer and BiLSTM did not reach statistical significance ($p = 0.098$), ST-Transformer's higher performance metrics for all test years suggest practical advantages for operational deployment.
This demonstrates the value of spatio-temporal modeling capabilities for predicting hypoxia in the Gulf of Mexico..
Future work will examine the transferability \cite{collier2018cactusnets} of these models to other costal locations where hypoxic conditions can occur. 
% We found that ST-Transformer consistently outperformed the other three models in predicting hypoxic events by achieving AUC-ROC values of 0.982-0.992.
% This demonstrates superior spatio-temporal modeling capabilities of ST-Transformer for predicting hypoxia in the Gulf of Mexico.

% \todo{McNemar's tests revealed statistical differences among model pairs.
% Most tests showed statistically significant differences $(p < 0.001)$ with large effect sizes.}
% Validation on 2022-2024 data demonstrates operational viability for real-world deployment.

% Current models lack sufficient temporal granularity for early warning systems.
% Our approach enables daily forecasting to support timely ecosystem management decisions.

%
%
%%%%%%%%%%%%%%%%%%%%%%%%%%%%%%%%%
%
%
% \section*{Acknowledgment}

% \bibliographystyle{IEEEtran}
\bibliographystyle{plainnat}
\bibliography{references}

\end{document}